\documentclass[letterpaper]{article}
\usepackage{natbib}
\usepackage{aaailike}
\usepackage{times}
\usepackage{helvet}
\usepackage{courier}
\usepackage[utf8]{inputenc}
\usepackage[T1]{fontenc}
\usepackage{amsmath,amssymb}
\usepackage{amsthm}
\usepackage{graphicx}
\usepackage{booktabs}
\usepackage{multirow}
\usepackage{array}
\usepackage{tabularx}
\usepackage{xcolor}
\usepackage{microtype}
\usepackage{tikz}
\usetikzlibrary{positioning,arrows.meta}
\usepackage{url}
\usepackage[skins,breakable]{tcolorbox}
\theoremstyle{definition}
\newtheorem*{definition}{Definition}

\definecolor{ourscol}{HTML}{D42300}
\definecolor{bbcol}{HTML}{1868B2}
\definecolor{cbcol}{HTML}{8048AA}
\definecolor{appcol}{HTML}{148843}

\newcommand{\sv}[1]{\textsc{\small #1}} 

\title{Faithful by Definition: Emotion Analysis\\
       via Natural Semantic Metalanguage Explications}
\author {
    Frank Xing\textsuperscript{*} and Erik Cambria\textsuperscript{\#}
}
\affiliations {
    \textsuperscript{*}University of Reading\ \ \ \ \textsuperscript{\#}Nanyang Technological University\\
    z.xing@henley.ac.uk\ \ \ \ cambria@ntu.edu.sg
}

\begin{document}
\maketitle

\begin{abstract}
Explanations for emotion classifiers are usually produced post hoc, with no guarantee that they reflect the computation behind the label. We present an explication interface for event-based emotion analysis. A parser maps the input text to an explication, a short script in the closed vocabulary of Natural Semantic Metalanguage organized into twelve typed slots, and a fixed decision list of rules transcribed from published semantic definitions computes the label from the explication alone. The faithfulness guarantee is therefore \emph{causal} and \emph{definitional}, while all empirical risk lives in the learned parser, which the per-line entailment interface makes auditable against the input. On crowd-sourced event descriptions, our fine-tuned parser reaches 0.33 accuracy and 0.48 selective accuracy on a small held-out set, suggesting that the interface trades insignificant accuracy difference to a black-box model for a verifiable, inspectable decision basis for first-person event-based emotion analysis. We also release EmoExpl-1200 with per-line verification metadata and the full rule set.
\end{abstract}

\section{Introduction}
Emotion analysis now informs content moderation, public-health screening, customer research, and the evaluation of conversational agents \citep{rajamanickam2020,ma2020empathetic}. In these settings a label alone is rarely sufficient; practitioners need to know why the system produced it, and regulators increasingly require the same. The dominant explanation formats do not meet this need. Post-hoc token attributions frequently disagree with the model's actual decision process \citep{jacovi2020}, and free-text rationales generated alongside an answer can rationalize the computation instead of reporting it \citep{lanham2023, madsen2024}. The field has responded mainly by measuring unfaithfulness more carefully, by using counterfactual conflicts, or by optimizing explanations toward faithfulness proxies; all routes are post hoc \citep{malandri2024} and leave the central guarantee missing.
\begin{figure}[t]
\centering
\begin{tcolorbox}[title = \textbf{frustration}, width=0.95\columnwidth]
\small{X feels something\\
sometimes a person thinks something like this:\\
I want to do something\\
I can't do this\\
because of this, this person feels something bad\\
X feels like this}
\end{tcolorbox}
\caption{A prototypical NSM explication that defines \emph{frustration} entirely in semantic primes \citep{wierzbicka1999}.}
\label{fig:nsm-frustration}
\end{figure}

\begin{figure*}[t]
\centering
\begin{tikzpicture}[
  >=Latex, node distance=6mm and 8mm, line width=0.5pt, font=\small,
  b/.style={draw, rounded corners, align=center, inner xsep=4pt, inner ysep=3pt, minimum height=12mm},
  stage/.style={b, minimum width=16mm, minimum height=15mm}]
\node[b, align=left] (t) {text:\\\itshape ``I got\\ \itshape the management\\ \itshape position''};
\node[stage, right=of t] (p) {\\[2.2ex]\textbf{parser} $f_\theta$\\(learned)};
\node[b, right=of p, align=left] (e) {\textbf{Emotion Explication Schema}\\[-1pt]\scriptsize\setlength{\tabcolsep}{2.5pt}\renewcommand{\arraystretch}{0.95}%
  \begin{tabular}{@{}l@{\;}l@{\quad}l@{\;}l@{}}
  EXPER. & \sv{i} & EVAL & \textbf{\sv{feel-good}}\\
  TRIGGER & \sv{someone-did} & E-TARGET & \textbf{\sv{self}}\\
  AGENCY & \textbf{\sv{i}} & KNOW & \sv{none}\\
  WANT & \sv{want} & O-KNOW & \sv{none}\\
  REALIZ. & \sv{happened} & BODY & \sv{no}\\
  TIME & \sv{before-now} & INTENS. & \sv{plain}\\
  \end{tabular}};
\node[stage, right=of e] (m) {\\[2.2ex]\textbf{mapper} $g$\\(fixed)};
\node[b, right=of m] (y) {emotion:\\[2pt] pride};
\node[b, below=12mm of e, minimum width={width("xEXPER.x\sv{i}xEVALx\textbf{\sv{feel-good}}x")}, align=center] (v)
  {\scriptsize NLI verifier: each line entailed by the text?};
\draw[->] (t) -- (p);
\draw[->] (p) -- (e);
\draw[->] (e) -- (m);
\draw[->] (m) -- (y);
\draw[->] ([xshift=-3mm]e.south) to[bend right=20] ([xshift=-3mm]v.north);
\draw[->] ([xshift=3mm]v.north) to[bend right=20] ([xshift=3mm]e.south);
\draw[->, dashed] (t.south) |- (v.west);
\node[below=1.5mm of m, font=\scriptsize, align=center] {\sv{feel-good}\,$\wedge$\,\sv{self}\\$\wedge$\,AGENCY{=}\sv{i}\\$\Rightarrow$ \textbf{pride}};
\begin{scope}[x=1mm, y=1mm, line width=0.5pt, shift={(p.north west)}]
  \foreach \aa in {-1.6,-3.2,-4.8}\foreach \bb in {-2.4,-4.0} \draw (2.4,\aa)--(5.2,\bb);
  \foreach \yy in {-1.6,-3.2,-4.8} \draw[fill=white] (2.4,\yy) circle (0.58);
  \foreach \yy in {-2.4,-4.0}      \draw[fill=white] (5.2,\yy) circle (0.58);
\end{scope}
\begin{scope}[x=1mm, y=1mm, line width=0.5pt, shift={(m.north west)}]
  \draw (2.2,-1.3) rectangle (5.5,-5.6);
  \draw (2.9,-2.3)--(4.3,-2.3);
  \draw (2.9,-3.3)--(4.3,-3.3);
  \draw (2.9,-4.3)--(4.3,-4.3);
  \draw[fill=white] (5.0,-4.6) circle (1.15);
  \draw[line width=0.8pt] (5.8,-5.4)--(6.8,-6.4);
\end{scope}
\end{tikzpicture}
\caption{The explication interface, shown on a corpus instance. The parser turns \textit{``I got the management position''} into the twelve-slot explication (centre); the fixed mapper fires the highest-priority rule whose conditions the explication satisfies, here \sv{feel-good} $\wedge$ EVAL-TARGET{=}\sv{self} $\wedge$ AGENCY{=}\sv{i} (the bold slots), and returns \textbf{pride}, a positive self-evaluation distinct from the broad joy route. Only the parser is learned, so causal faithfulness of the mapper step holds by construction; each explication line is a proposition the verifier checks against the input text (dashed edge), with claims flowing from the explication to the verifier and entailment judgments flowing back along the two parallel arcs.}
\label{fig:interface}
\end{figure*}
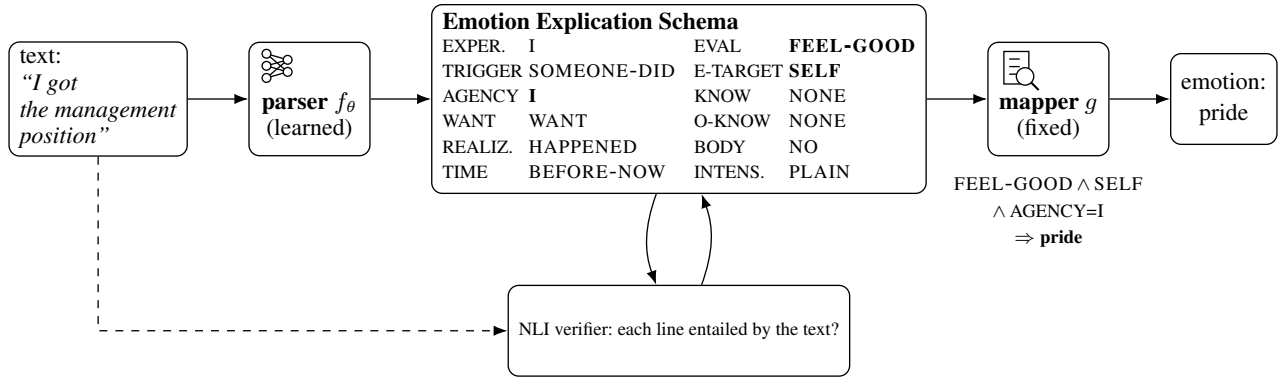

This paper pursues a \emph{constructive alternative} for one task family: the prediction pathway itself serves as the explanation. We operationalize Natural Semantic Metalanguage (NSM), a linguistic-semantics program that defines word meanings through a closed set of semantic primes (Section~2), as a two-segment \emph{definitional pathway} $\hat y = g(f_\theta(x))$. A parser $f_\theta$ maps text to an explication in a twelve-slot schema over the prime vocabulary, and a transparent mapper $g$, derived from the published definitions, computes the label from the explication and nothing else. The second segment is fixed by semantic theory; the first segment, which targets a cognitively and sensorily more primitive representation, is the only part that must be learned (Figure~\ref{fig:interface}).

Three developments make the design feasible now. LLMs can generate explications that respect the prime vocabulary \citep{deepnsm2025}; constrained decoding enforces the closed vocabulary so legality is a guaranteed decoder property; mature natural language inference (NLI) models make per-proposition verification practical at corpus scale; and appraisal-annotated corpora supply event descriptions with the cognitive granularity the schema requires \citep{troiano2023}.

The paper makes three contributions. \emph{First}, we introduce the Emotion Explication Schema (EES), a twelve-slot closed-vocabulary representation for emotional construals, and a definitional rule mapper over thirteen emotions. We are, to our knowledge, the first to combine an NSM-grounded closed vocabulary with instance-level emotion classification and a per-line verification interface, with three scoped properties: (i)~the explication-to-label segment is causally faithful by construction (structural transparency: the label is computed from the explication alone); (ii)~the vocabulary is a closed, independently motivated prime set; and (iii)~the text-to-explication segment is verifiable line by line through entailment. We scope all claims to first-person, event-based emotions (Section~2). \emph{Second}, we release EmoExpl-1200, an instance-level explication-annotated corpus over crowd-sourced event descriptions derived from crowd-enVent \citep{troiano2023}, with per-line verification metadata.\footnote{Code, the EmoExpl-1200 corpus, and the rule set are available at \url{https://github.com/fxing79/ebm}.} \emph{Third}, we document a replicable methodology for revising definitional rules under regression safeguards, including a case in which textual evidence forced a revision to be rolled back.

\section{Background and Notation}
\paragraph{Natural Semantic Metalanguage.}
NSM analyzes meaning through roughly sixty-five \emph{semantic primes}: simple, cross-linguistically attested concepts such as \sv{I}, \sv{someone}, \sv{do}, \sv{happen}, \sv{want}, \sv{know}, \sv{feel}, \sv{good}, and \sv{bad} \citep{wierzbicka1996, goddard2014}. Primes are posited as indefinable; every other word sense is defined by an \emph{explication}, a short prototypical script in primes. For emotions this yields a paraphrase of the prototypical eliciting situation. A pride-like state reads, in part, ``I did something; I think this is good; I think something good about myself; I feel something good because of this'' \citep{wierzbicka1999}. Crucially, the explication \emph{is} the definition of the concept, fixed by the theory independently of any classification task. For instance, the explication of \emph{frustration} reads, in part, ``I wanted to do something; I now know I cannot do this; because of this I feel something bad'' \citep{wierzbicka1999}, which our schema records as WANT=\sv{want}, REALIZATION=\sv{not-can}, EVALUATION=\sv{feel-bad} (Figure~\ref{fig:nsm-frustration}). NSM has been applied at scale to emotion vocabularies organized by elicitor category and by cognitive template \citep{wierzbicka1999}; Appendix~\ref{app:nsm-bkgd} reproduces these classifications for reference.

\paragraph{Notation.}
We write $x\in\mathcal{X}$ for an input text, $c\in\mathcal{C}$ for an explication, and $y\in\mathcal{Y}$ for an emotion label, with $\mathcal{Y}$ the thirteen categories plus a reserved $\textsc{abstain}$ symbol. The system is the composition
\begin{equation}
\hat{y} \;=\; g\big(f_\theta(x)\big), \qquad f_\theta:\mathcal{X}\!\to\!\mathcal{C},\quad g:\mathcal{C}\!\to\!\mathcal{Y},
\label{eq:pipeline}
\end{equation}
where the \emph{parser} $f_\theta$ is the only learned component and the \emph{mapper} $g$ is a fixed function specified in advance. We call this a \emph{definitional pathway}: all empirical risk lives in $f_\theta$, while $g$ is determined by the NSM definitions and never sees a label during training.

\paragraph{The Emotion Explication Schema.}
We discretize the prime configurations relevant to event-based emotion into a finite product space.
\begin{definition}[EES]
The Emotion Explication Schema is the typed product $\mathcal{C}=\prod_{s=1}^{12} V_s$ over twelve slots with finite value sets $V_s$ (Table~\ref{tab:schema}), partitioned into event structure, appraisal, and expression. A renderer $r:\mathcal{C}\to 2^{\mathcal{P}}$ expands an assignment into a set of prime-vocabulary propositions $\mathcal{P}$, one per active slot. An assignment is \emph{legal} iff every value lies in its $V_s$ and every rendered line parses under the prime grammar.
\end{definition}
The schema is the engineering object the paper studies; it indexes prime configurations and does not extend the prime inventory. An instance pairs the twelve-slot assignment with a \emph{residue} field, which records construals the primes cannot express and stays empty under full coverage, and a free-text \emph{notes} field.

\begin{table}[t]
\centering
\small
\setlength{\tabcolsep}{4pt}
\begin{tabular}{@{}llp{4.9cm}@{}}
\toprule
Group & Slot & Value set $V_s$ \\
\midrule
\multirow{5}{*}{\rotatebox{90}{event}}
 & EXPERIENCER & \sv{i}, \sv{someone}, \sv{people} \\
 & TRIGGER & \sv{something-happened}, \sv{someone-did-something}, \sv{nothing}, \sv{none} \\
 & AGENCY & \sv{i}, \sv{someone-else}, \sv{no-one}, \sv{none} \\
 & REALIZATION & \sv{happened}, \sv{not-happened}, \sv{can}, \sv{not-can}, \sv{maybe}, \sv{none} \\
 & TIME-DIR.\ & \sv{before-now}, \sv{now}, \sv{after-now} \\
\midrule
\multirow{4}{*}{\rotatebox{90}{appraisal}}
 & WANT & \sv{want}, \sv{not-want}, \sv{none} \\
 & EVALUATION & \sv{feel-good}, \sv{feel-bad}, \sv{neither} \\
 & EVAL-TARGET & \sv{self}, \sv{other}, \sv{event}, \sv{object}, \sv{none} \\
 & KNOWLEDGE & \sv{know}, \sv{not-know}, \sv{maybe-know}, \sv{none} \\
\midrule
\multirow{3}{*}{\rotatebox{90}{expr.}}
 & OTHERS-KNOW & \sv{can-know}, \sv{none} \\
 & BODY & \sv{yes}, \sv{no} \\
 & INTENSITY & \sv{very}, \sv{plain}, \sv{small} \\
\bottomrule
\end{tabular}
\caption{The twelve-slot EES and its closed value sets. The full Cartesian space has $|\mathcal{C}|\approx 1.9{\times}10^{6}$ legal assignments, of which the mapper names thirteen emotion regions and can elaborate further. The schema targets first-person, event-based emotions; nested time, relational, and aspectual construals fall outside it.}
\label{tab:schema}
\end{table}

\section{Related Work}
\paragraph{Faithfulness, disambiguated.}
``Faithfulness'' labels several distinct questions, and our guarantee concerns exactly one. \emph{Explanation faithfulness} asks whether an explanation reflects the computation behind a prediction \citep{jacovi2020}; work here either measures the property \citep{walkthetalk2025,madsen2024,yeo2025,phicct2025} or optimizes toward it \citep{srnle2025,faithlm2024}. We instead establish it by construction for one segment. The other senses are orthogonal: \emph{contextual} and \emph{situated faithfulness} concern grounding in supplied documents \citep{faitheval2025,canoe2026,cognibench2025,totrust2025}; \emph{reasoning faithfulness} concerns whether stated steps support an answer \citep{fidelis2024,breakingchain2026}; \emph{confidence faithfulness} concerns verbalized confidence \citep{metafaith2025,confgap2026}; \emph{reconstruction fidelity} concerns sparse decompositions \citep{mxd2025,intrinsic2026}; and one further sense concerns fidelity to one's own judgment under pressure \citep{afice2025}. Inside bottleneck models, \emph{structural} faithfulness (the label is computable only from the intermediate layer) is cheap, while \emph{input} faithfulness (that layer truly describes the input) is the hard part. Concept-bottleneck LLMs inherit the first and neither guarantee nor verify the second \citep{cbllm2025}; our design assigns the structural part to the definitional mapper and routes all input-faithfulness risk to the parser, where the per-line verifier measures it.

\paragraph{Closed-vocabulary neurosymbolic emotion analysis, appraisal theory, self-conscious emotions, and NSM in NLP.}
A neurosymbolic tradition already pairs a closed primitive vocabulary with neural models \citep{senticnet9}; The differences here are: (1) our intermediate layer is a \emph{propositional script with a grammar} rather than a flat polarity lexicon, and (2) we parse at the \emph{instance} level rather than concepts lookup. Generic concept-bottleneck LLMs \citep{koh2020,cbllm2025}, however, satisfy neither. Appraisal theories \citep{occ1988,scherer2009,smith1985,troiano2023} are already propositional and rule-based; the difference is the \emph{linguistic} motivation of our vocabulary and the per-line verifiability of discrete propositions. Shame and guilt are distinguished in psychology by global self-blame versus specific behavior-blame \citep{tangney2002,tracy2004}; our mapper separates them with a single prime (OTHERS-KNOW versus AGENCY), a deliberate simplification we adopt for transcribability. Automated explication at the word-sense level is demonstrated by \citet{deepnsm2025} and motivates our parser; mechanistic work finds appraisal concepts steerable in LLM emotion inference, with agency steering converting sadness into guilt \citep{tak2025}, which motivates but does not validate our symbolic slot-flips. A probing experiment that tests whether the parser's hidden states encode the slots is planned as future work.

\section{Method}
The interface has three components over EES: a learned parser $f_\theta$, a fixed mapper $g$, and a per-line verifier NLI.

\subsection{The parser $f_\theta$}
The parser maps text to a twelve-slot assignment. Under schema-constrained decoding $f_\theta(x)\in\mathcal{C}$ for every $x$ and $\theta$, so legality is one by configuration; we therefore report \emph{measured} legality under free decoding (Section~6), where it is informative. We study a zero-shot prompted parser and fine-tuned variants (Section~5). All empirical risk concentrates in $f_\theta$; the verifier of Section~4.3 audits it line by line. A set of written \emph{construal conventions} governs how an annotator resolves slots that the text underdetermines (for example, a completed event narrated in the present tense takes TIME-DIRECTION=\sv{before-now}; a slot leaves its default only on explicit textual support). These conventions are part of the annotation protocol, not of the guarantee.

\subsection{The definitional mapper $g$}
\begin{definition}[Mapper]
$g$ is a priority-ordered decision list of thirteen rules $(e_k,\phi_k)_{k=1}^{13}$, where $e_k\in\mathcal{Y}$ is an emotion and $\phi_k$ is a conjunction of slot-value conditions transcribed from the published NSM explication of $e_k$. For an explication $c$, $g(c)=e_k$ for the smallest $k$ with $c\models\phi_k$, and $g(c)=\textsc{abstain}$ if no rule fires.
\end{definition}
The order is specific-before-general, so a particular positive emotion shadows the broad joy rule, and abstention rate is itself reported. Table~\ref{tab:rules} lists representative rules; the full thirteen-rule list is in Appendix~\ref{app:nsm-bkgd}. The self-conscious pair is instructive: shame and guilt share \sv{feel-bad} and a self-directed evaluation and are separated by a single prime, whether others can know (\sv{others-know}) versus whether the experiencer is the agent (\sv{agency}).

\begin{table}[t]
\centering
\small
\setlength{\tabcolsep}{4pt}
\begin{tabular}{@{}llp{5.7cm}@{}}
\toprule
$k$ & $e_k$ & condition $\phi_k$ \\
\midrule
$\cdots$ & relief & WANT=\sv{not-want}, REALIZATION=\sv{not-happened}, EVALUATION=\sv{feel-good} \\
$\cdots$ & shame & EVALUATION=\sv{feel-bad}, EVAL-TARGET=\sv{self}, OTHERS-KNOW=\sv{can-know} \\
$\cdots$ & guilt & EVALUATION=\sv{feel-bad}, EVAL-TARGET=\sv{self}, AGENCY=\sv{i} \\
$\cdots$ & pride & EVALUATION=\sv{feel-good}, EVAL-TARGET=\sv{self}, AGENCY=\sv{i} \\
$\cdots$ & trust & EVALUATION=\sv{feel-good}, EVAL-TARGET=\sv{other}, KNOWLEDGE=\sv{know} \\
$\cdots$ & anger & EVALUATION=\sv{feel-bad}, AGENCY=\sv{someone-else}, EVAL-TARGET=\sv{other} \\
last & joy & EVALUATION=\sv{feel-good}, TRIGGER$\in\{$\sv{something-happened}, \sv{someone-did}$\}$ \\
\bottomrule
\end{tabular}
\caption{Representative rules from $g$, in priority order. Each $\phi_k$ is transcribed from a published explication; no condition is learned. The full list is in Appendix~\ref{app:nsm-bkgd}.}
\label{tab:rules}
\end{table}

Because $g$ reads only $c$, two properties hold by construction. The label is invariant to anything not encoded in $c$; and for any single-slot edit $c\!\to\!c'$ the label changes exactly as the firing conditions dictate, so the Causal Consistency Rate equals $1.0$ as an identity. This is true of \emph{any} deterministic function of $c$, so it is not by itself an XAI contribution. The scientific risk lives entirely in whether $c$ faithfully represents $x$, which the parser must get right and the verifier measures (Section~6). Figure~\ref{fig:flip} illustrates the identity: editing \sv{others-know} alone moves a guilt explication into the higher-priority shame rule.

\begin{figure}[t]
\centering
\small
\setlength{\tabcolsep}{4pt}
\begin{tabular}{@{}lcc@{}}
\toprule
slot & explication $c$ & edit $c'$ \\
\midrule
EVALUATION  & \sv{feel-bad} & \sv{feel-bad} \\
EVAL-TARGET & \sv{self}     & \sv{self} \\
AGENCY      & \sv{i}        & \sv{i} \\
OTHERS-KNOW & \sv{none}     & \textbf{\sv{can-know}} \\
\midrule
$g(\cdot)$  & \textbf{guilt} & \textbf{shame} \\
\bottomrule
\end{tabular}
\caption{A single-slot counterfactual. Flipping OTHERS-KNOW satisfies the higher-priority shame rule, so $g$ relabels guilt$\to$shame exactly as the definitions prescribe.}
\label{fig:flip}
\end{figure}

\subsection{Per-line verification and rule revision}
The renderer turns $c$ into discrete propositions, and a natural language inference model scores the entailment of each proposition by the input text $x$. The per-item verification score is the fraction of lines entailed above a threshold; it is an audit channel, calibrated against human entailment on a held-out subset before any gating use. Conventions and rules stay frozen between revisions; a revision passes only through an adjudication round, and every change runs a regression suite of thirteen canonical self-checks, a fifty-two-item pilot, and the slot-flip consistency test. One revision was rolled back when held-out evidence showed it stole visceral grief into the disgust route (Section~6); this episode is the basis of our replicability claim.

\section{Experimental Setup}
\paragraph{Corpus and splits.}
The corpus draws on crowd-enVent \citep{troiano2023}, where each writer described an episode for a prompted emotion; the prompted emotion is the reference label, whose imperfection Section~6 quantifies. We annotate $1{,}200$ items into EmoExpl-1200 with per-line verification metadata, and draw a $156$-item stratified subset (twelve per emotion) for two parallel passes: an \emph{LLM-annotator} pass in which one of the most powerful models (Claude~Fable~5) applies the written guidelines, and a \emph{gold} pass by two trained human annotators applying the same guidelines. Because this annotator and the parser are both LLMs and may share construal biases, we treat the LLM-annotator pass as weak supervision, never as inter-annotator agreement; a cross-family pass with a non-Claude annotator (Llama-3.3-70B-Versatile via Groq) on the same $156$ items returns the same slot reliability pattern at a lower scalar (mean $\alpha\!=\!0.429$ vs.\ within-Claude $0.491$), confirming the model-agreement reading. The held-out evaluation set is $36$ items drawn from the gold/silver core with a fixed seed; the leakage-free training pool is the remaining gold+silver explications ($206$ clean pairs, or $1{,}164$ with weak parser-preannotation augmentation). Table~\ref{tab:tiers} lists every evidence tier.

\begin{table}[t]
\centering\small
\begin{tabular}{llr}
\toprule
Tier & Source & $n$ \\
\midrule
P & author pilot & 52 \\
A & automatic metrics, full batch & 1{,}200 \\
S & silver (Claude~Fable~5, guidelines) & 156 \\
H & human dual annotation & 99 \\
gold & adjudicated human gold & 98 \\
eval & held-out routing set & 36 \\
\bottomrule
\end{tabular}
\caption{Evidence tiers and sizes. Tier A reports label-free automatic metrics, S model agreement, H inter-annotator reliability.}
\label{tab:tiers}
\end{table}

\paragraph{Parser configurations.}
The zero-shot parser is a prompted instruct model with constrained decoding. Fine-tuned parsers apply LoRA to \mbox{Llama-3.2-1B}, \mbox{Llama-3.2-3B}, and \mbox{Llama-3.1-8B}: rank $16$ ($\alpha{=}32$) for 1B and $32$ ($\alpha{=}64$) for 3B and 8B, dropout $0.05$, on attention and MLP projections; learning rates $2/1.5/1{\times}10^{-4}$; $3$--$5$ epochs; effective batch $16$; 8B uses 4-bit NF4 QLoRA. The 3B and 8B runs add the weak-augmentation pool. Each scale trains and is evaluated on a single A100 in well under an hour. At inference we report results under free decoding to make the learned-legality claim falsifiable.

\paragraph{Baselines.}
We compare four unconstrained families on the same $36$-item held-out set. (1)~Black-box classifiers: sequence-classification heads on the same Llama-3.2-1B/3.2-3B/3.1-8B backbones (LoRA, identical split), and a RoBERTa-large \citep{roberta} classifier trained on the full corpus as an upper bound. (2)Appraisal pipeline: a logistic mapping from the twenty-one gold crowd-enVent appraisal dimensions \citep{troiano2023} to emotion, trained on the corpus minus eval; it is given oracle appraisal at test time. (3) Concept-bottleneck LLM \citep{cbllm2025}; (4) a SenticNet 9 lexicon baseline \citep{senticnet9}, mean/max/min of four affective primitives plus polarity and Plutchik category counts, fed into a logistic head. The verifier is a DeBERTa-v3-large NLI model.

\paragraph{Metrics.}
Accuracy against the prompted emotion, abstention rate, selective accuracy on routed items~\citep{El-Yaniv2010}; explication legality; Krippendorff's $\alpha$ for reliability; the Causal Consistency Rate for $g$; and the per-line verification score. Each figure carries one tier label of Table~\ref{tab:tiers}.

\section{Results and Analysis}
\label{sec:results}
Each figure carries its evidence source. Automatic metrics (A) cover the full $1{,}200$-item batch. Silver figures (S) come from the $156$-item subset and report model agreement, never IAA. Gold figures (H) come from the human dual-annotation pass and its adjudication. Accuracy means agreement with the prompted emotion.

\subsection{Main classification}
\begin{figure}[t]
\centering
\includegraphics[width=\columnwidth]{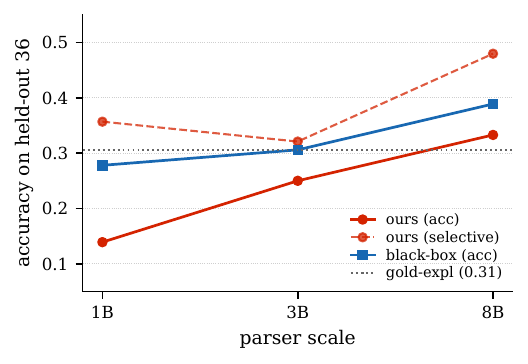}
\caption{Accuracy against parser scale on the held-out set. The fine-tuned parser (red) improves with scale; a same-supervision black-box classifier (blue) is higher in raw accuracy but the differences to ours diminish with scale; both are not statistically distinguishable from gold-explications.}
\label{fig:scale}
\end{figure}
The zero-shot parser with constrained decoding reaches $38.3\%$ accuracy at $13.8\%$ abstention on the full batch (selective $44.4\%$, legality $100\%$, A), and $36.5\%$ on the $156$-item subset. Fine-tuning sharpens routing on the held-out $36$ items (Table~\ref{tab:e2main}, Figure~\ref{fig:scale}): the 8B parser routes twenty-five items at $0.480$ selective accuracy and $0.333$ overall, abstaining on eleven, and all three scales hold legality at $100\%$ under free decoding, so the parser learns to stay legal without a decoding constraint. Routing the human gold explications through the same mapper yields $0.306$ ($0.458$ selective); this is the gold explication routed through $g$, not an upper bound. On raw accuracy the oracle-appraisal pipeline is numerically highest ($0.361$), the same-scale black box next ($0.389$), and our parser $0.333$; the concept-bottleneck LLM reaches $0.222$ and SenticNet 9 reaches $0.250$. We do \emph{not} claim an accuracy win. At $n{=}36$ the exact binomial intervals on all of these overlap (Appendix~\ref{app:mapper-details}, Table~\ref{tab:ci}), so no pairwise difference is statistically resolvable; the interface's distinguishing property is the verifiable decision basis (and a higher selective accuracy, $0.480$, on the items it routes).

\begin{table}[t]
\centering
\setlength{\tabcolsep}{4pt}
\small
\resizebox{\columnwidth}{!}{%
\begin{tabular}{lcccc}
\toprule
System (held-out, $n{=}36$) & Routed & Sel.\ acc & Acc & Legality \\
\midrule
\textbf{Fine-tuned 1B (ours)} & 14/36 & 0.357 & 0.139 & 1.00 \\
\textbf{Fine-tuned 3B (ours)} & 28/36 & 0.321 & 0.250 & 1.00 \\
\textbf{Fine-tuned 8B (ours)} & 25/36 & 0.480 & 0.333 & 1.00 \\
Gold explication (routed)   & 24/36 & 0.458 & 0.306 & 1.00 \\
\midrule
Black-box 1B                & 36/36 & 0.278 & 0.278 & n/a \\
Black-box 3B                & 36/36 & 0.306 & 0.306 & n/a \\
Black-box 8B                & 36/36 & 0.389 & 0.389 & n/a \\
Black-box RoBERTa$^{\dagger}$ & 36/36 & 0.111 & 0.111 & n/a \\
Appraisal$^{\dagger}$       & 36/36 & 0.361 & 0.361 & n/a \\
Concept-bottleneck LLM      & 36/36 & 0.222 & 0.222 & n/a \\
SenticNet 9 lexicon         & 36/36 & 0.250 & 0.250 & n/a \\
\bottomrule
\end{tabular}}
\caption{Main results on the held-out $36$-item set. Fine-tuned parsers (ours) route through the mapper and may abstain. Direct classifiers never abstain. $\dagger$ marks systems given extra information (RoBERTa trained on the full corpus; appraisal given oracle ratings).}
\label{tab:e2main}
\end{table}

\subsection{Guideline executability and reliability}
\begin{table}[t]
\centering\small
\begin{tabular}{lccc}
\toprule
Annotation ($156$-item subset) & Acc & Abstain & Sel.\ acc \\
\midrule
\textbf{LLM-annotator} (model) & 0.538 & 0.064 & 0.575 \\
Parser (zero-shot) & 0.365 & 0.115 & 0.413 \\
\bottomrule
\end{tabular}
\caption{Guideline executability (S). Routed through the same mapper, silver outscores the zero-shot parser by $17$ points. The four-cell split is $44$ both correct, $40$ silver-only, $13$ parser-only, $59$ both wrong.}
\label{tab:exec}
\end{table}
\begin{figure*}[t]
\centering
\includegraphics[width=0.9\textwidth]{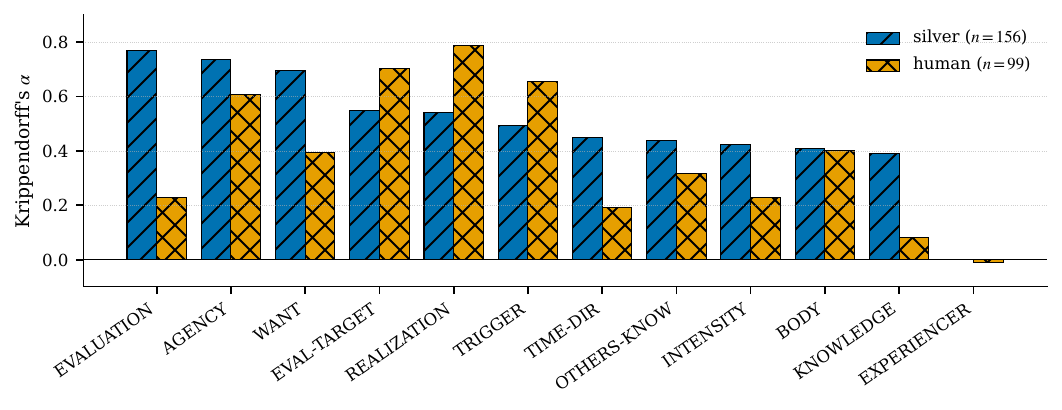}
\caption{Per-slot Krippendorff's $\alpha$: silver model agreement (blue, $n{=}156$) and the first human pass on the adversarial stratum (yellow, $n{=}99$). Event-structure and core appraisal slots reach $\alpha\ge 0.67$; KNOWLEDGE and the variance-degenerate EXPERIENCER collapse. Exact numbers appear in Appendix~\ref{app:perslot}.}
\label{fig:slotalpha}
\end{figure*}
The LLM-annotator explications reach $53.8\%$ accuracy at $6.4\%$ abstention (selective $57.5\%$) against the parser's $36.5\%$/$11.5\%$/$41.3\%$ (Table~\ref{tab:exec}, S). The $17$-point gap reflects partly prompting and training differences, not just guideline executability: the cross-family pass on the same $156$ items, replacing Claude~Fable~5 with Llama-3.3-70B-Versatile, returns mean per-slot $\alpha=0.429$ (vs.\ $0.491$ within-Claude) and preserves the slot ordering, so silver is model agreement rather than IAA \citep{xing2020coling}. Per-slot, raw agreement averages $0.803$ across the twelve slots while Krippendorff's $\alpha$ averages $0.491$, or $0.536$ excluding the variance-degenerate EXPERIENCER (Figure~\ref{fig:slotalpha}, Appendix~\ref{app:perslot}). The three slots that carry most routing decisions, EVALUATION, WANT, and AGENCY, hold $\alpha\ge 0.68$, while five fall below $0.45$ and define the calibration priorities. Divergences are directional: the parser reads human-caused events as agentless happenings ($34$ items on TRIGGER), marks present-tense retellings as NOW ($24$ on TIME-DIRECTION), and over-specifies diffuse evaluations to persons ($21$ on EVAL-TARGET).

\paragraph{Human pass and adjudication.}
Two trained annotators independently labeled $99$ items from the highest-priority review stratum. Raw agreement averages $0.809$ while $\alpha$ averages $0.384$, replicating the inflation pattern. Determinacy splits by slot type: four event-structure slots hold $\alpha$ between $0.67$ and $0.78$, while five collapse below $0.30$ (KNOWLEDGE $0.04$, TIME-DIRECTION $0.19$, INTENSITY $0.23$, EVALUATION $0.25$, plus degenerate EXPERIENCER). Conflicts are one-directional. Routed through the mapper, the human explications abstain on roughly $80\%$ of items (the two annotators agree on abstaining for $64$ of $99$), exposing a routing frontier rather than estimating accuracy. The interface pre-filled slots from parser predictions, and each annotator kept those defaults on $0.85$ and $0.89$ of slots against $0.81$ between themselves, so the first-batch figure measures \emph{anchored verification}. A convention vote resolved the directional conflicts into a gold standard of $98$ items under one principle: a slot takes a marked value only on explicit textual evidence. EVALUATION took a calibrated exception: a bare negative life event (e.g.\ divorce) maps to FEEL-BAD; we acknowledge this encodes a mild world-knowledge prior. The adjudicated gold abstains on $85$ of $98$ items, confirming the routing frontier.

\subsection{Distributional evaluation}
Per-emotion accuracy ranges from joy ($0.70$) down to trust ($0.09$); the prominent off-diagonal flows in Table~\ref{tab:offdiag} localize the failures. The largest single source is a slot-rule interaction: fourteen silver explications carry KNOWLEDGE=\sv{not-know} from discovery framings, thirteen route to surprise, and the slot has the lowest $\alpha$ in the corpus. The adjudication's evidence gate on KNOWLEDGE targets exactly this. A second localization concerns the mapper: $28.5\%$ of parser explications satisfy more than one rule, resolved by priority; most collisions are specific-over-general by design, and one reflects a true overlap, shame ahead of guilt on $66$ items. On the adjudicated gold the tie rate is zero. Typed abstention, priority-order sensitivity, and a distributional evaluation that scores the mapper as a 13-way distribution over fired rules (priority-decay strictly dominates single-winner, lifting top-3 by six points at $n{=}1{,}200$) are reported in Appendix~\ref{app:mapper-details}.

\begin{figure}[t]
\centering
\includegraphics[width=\columnwidth]{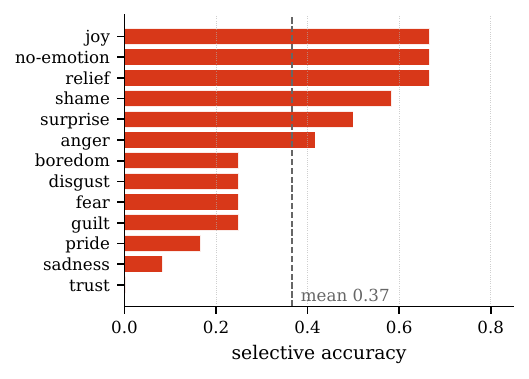}
\caption{Selective accuracy by prompted emotion (A, $n{=}1200$). Recovery is strong for joy, anger, and no-emotion and weak for trust and fear; the dashed line is the overall mean ($0.38$).}
\label{fig:peremotion}
\end{figure}
\begin{table}[t]
\centering\small
\setlength{\tabcolsep}{4.5pt}
\begin{tabular}{llrr}
\toprule
gold $i$ & routed $j$ & $n$ & \% of $i$ \\
\midrule
trust    & joy        & 46 & 50.5 \\
pride    & joy        & 44 & 47.3 \\
guilt    & shame      & 34 & 36.6 \\
surprise & joy        & 34 & 36.6 \\
relief   & joy        & 30 & 32.6 \\
shame    & guilt      & 25 & 27.5 \\
disgust  & anger      & 23 & 24.7 \\
disgust  & surprise   & 22 & 23.7 \\
boredom  & no-emotion & 17 & 18.5 \\
fear     & surprise   & 16 & 17.2 \\
\bottomrule
\end{tabular}
\caption{Prominent off-diagonal routing flows on the full batch (A, $n{=}1200$); the complete row-normalized confusion matrix is Figure~\ref{fig:confusion} in Appendix~\ref{app:verifier}. Positive emotions collapse toward the broad joy route and the self-conscious pair swaps in both directions; both patterns trace to the lowest-reliability slots.}
\label{tab:offdiag}
\end{table}

\subsection{Verification and counterfactual analysis}
The entailment verifier assigns a mean per-item score of $0.201$, with $487$ of $1{,}200$ items at zero (Appendix~\ref{app:verifier}, Figure~\ref{fig:nli}), while selective accuracy stays at $44.4\%$ across this range (A). Verification and routing respond to different signals: the verifier is conservative on prime-style minimal sentences, a known cost of the closed vocabulary, while routing depends only on slot values. We therefore treat the score as an audit channel and defer any gating decision to calibration. The Causal Consistency Rate equals $1.0$ as an identity (Section~4); single-slot flips reproduce the documented contrasts, including guilt$\to$shame through OTHERS-KNOW alone (Figure~\ref{fig:flip}). To probe whether the contrast survives at the text level, we built a $42$-pair counterfactual benchmark over six slot-edit families (Table~\ref{tab:cf}) where each pair $(x, x')$ rewrites the input to flip a designated slot. The parser flips the intended slot on $76\%$ of pairs and the definitional pathway carries the contrast to the expected emotion on $36\%$. Clean-flip is near zero by design: a real text rewrite shifts surface signals across multiple slots, so the parser's incidental updates are expected, and the structural-transparency property still localizes the mapper-side causal effect under $g$. Per-family rates track the slot reliabilities of Figure~\ref{fig:slotalpha}: the high-$\alpha$ multi-slot relief recipe (F2) leads at label-flip $0.57$; KNOWLEDGE (F3) and the EVAL-TARGET-only swap into a residue region (F6) bottom at $0.14$.
\begin{table}[t]
\centering\small
\setlength{\tabcolsep}{4pt}
\begin{tabular}{lcrrrr}
\toprule
Family & slot flip & $n$ & slot & label & causal \\
\midrule
F1  & OTHERS-KNOW          & 7 & 0.43 & 0.29 & 0.00 \\
F2  & WANT + REALIZATION   & 7 & 1.00 & 0.57 & 0.14 \\
F3  & KNOWLEDGE            & 7 & 0.86 & 0.14 & 0.00 \\
F4  & AGENCY + EVAL-TARGET & 7 & 0.71 & 0.43 & 0.00 \\
F5  & TIME + REALIZATION   & 7 & 1.00 & 0.57 & 0.00 \\
F6  & EVAL-TARGET only     & 7 & 0.57 & 0.14 & 0.00 \\
\midrule
\textbf{all} & --- & \textbf{42} & \textbf{0.76} & \textbf{0.36} & \textbf{0.02} \\
\bottomrule
\end{tabular}
\caption{Counterfactual benchmark. \emph{Slot} and \emph{label} are the rates at which the parser flips the targeted slot and routes both endpoints to the expected emotions; \emph{causal} requires both, with clean-flip gating it to near zero by design.}
\label{tab:cf}
\end{table}

\subsection{Inherent disagreement and ground truth}
The silver annotator flagged $32$ of $156$ items ($20.5\%$) as admitting two defensible construals, $14$ ($9.0\%$) as unrecoverable under masking, and three as vicarious; these rates bound end-to-end accuracy on this corpus well below $100\%$. A low $\alpha$ may raise a foundational objection: if trained annotators disagree this much, perhaps the items have no determinate label \citep{plank2022, pavlick2019, uma2021}. Two replies are decisive. First, the corpus places ground truth at the source (the writer's self-report about an experienced episode), so disagreement among readers measures recoverability from a degraded text rather than indeterminacy. Second, determinacy is slot-heterogeneous: the median disagreeing pair differs on two of twelve propositions, so a label conflict such as shame versus guilt decomposes into ten agreed and one or two contested propositions, a distinction a black-box label cannot register. Where two construals survive adjudication, the mapper can emit both with the pivotal slot that separates them \citep{rottger2022, uma2021}.

\section{Discussion}
\paragraph{Scope of the guarantee.}
The faithfulness guarantee covers the explication-to-label segment only; the parser remains learned and carries all empirical risk. The phrase ``faithful by definition'' must always be read with the segment qualifier attached. Verification, not the guarantee, does the empirical work, and our verifier is conservative on prime-vocabulary sentences, which depresses raw scores; we therefore report it as an audit signal to be calibrated, never as a gate.

\paragraph{How to read the reliability numbers.}
Two biases shape the first human pass and we flag both. The stratum is adversarial by selection, which lower-bounds the agreement of a random batch; the interface also pre-filled slots from parser predictions, anchoring annotators and inflating agreement above blind annotation. Because one bias deflates and the other inflates, the per-slot determinacy \emph{pattern}, stable across both, carries the interpretive load rather than the scalar $\alpha$.

\paragraph{Theory dependence and expressivity.}
The contested universality claims of NSM are not load-bearing here: we use only the closed vocabulary, the propositional format, and the availability of independently motivated definitions. The frozen schema cannot express every construal: remembered fear needs nested time, vicarious pride a relation slot, terminated relief an aspectual distinction; we flag such instances, report the residue rate, and restrict the headline claims to event-based emotions.

\paragraph{Future work.}
The current evidence base is constrained by annotator hours, compute, and API budget; what follows is the planned, not the completed, programme. (i)~A blind, random-stratum human pass would replace the present anchored, adversarial estimate. (ii)~Parser probing would test whether slot decisions are driven by the documented features or by surface artefacts. (iii)~Scaled-up annotation would tighten the held-out interval beyond $n{=}36$. (iv)~Calibrating the verifier against human entailment judgments turns the audit channel into a usable gate. (v)~A controlled simulatability study \citep{lyu2024} would test verifiability head-on. (vi)~A cross-lingual evaluation would test whether slot-level construals transfer across the languages in which the primes are attested.

\section{Conclusion}
We present an explication interface for event-based emotion analysis, in which faithfulness is relocated from a post-hoc measurement target to a structural property of the pipeline architecture. On crowd-sourced event descriptions, we fine-tuned an 8B parser that reaches 0.33 accuracy and 0.48 selective accuracy on the held-out set (n=36), statistically indistinguisable from a same-scale black-box emotion predictor (0.39 acc. and \textasciitilde{}0.50 sel. acc.) or human reconstruction (0.39 acc. and 0.48 sel. acc.). 

Without compromising accuracy, however, our interface makes the analytical steps of emotion auditable line by line. We release the schema, the mapper, and EmoExpl-1200 with per-line verification metadata, together with the rule-revision protocol that produced them. 
In future work, probing the parser's hidden states for slot encoding would test whether the interface reflects the model's internal working mechanism.
\bibliography{bib/refs}

\appendix
\section{Schema Reference and Full Rule Set}
\label{app:nsm-bkgd}
This appendix collects the rest of the interface specification: the NSM source classifications the schema discretizes and the complete thirteen-rule mapper. The NSM literature classifies emotion concepts by elicitor category (Table~\ref{tbl:ec}) and by cognitive template (Table~\ref{tbl:tem}), and gives short prime-only explications for each concept (Figure~\ref{fig:nsm-frustration} in the main text is one). Our twelve-slot EES discretizes these prime configurations rather than replacing the prime inventory; Table~\ref{tab:rules-full} then lists the complete mapper that reads the resulting slots.

\begin{table*}[ht]
\caption{Emotion concepts analyzed with the NSM approach by~\citet{wierzbicka1999}.}
\label{tbl:ec}
\centering
\begin{tabularx}{\textwidth}{lp{12.6cm}}
\toprule
Category & Emotion concepts \\
\midrule
Bad things happening & sad, unhappy, distressed, upset, sorrow, sorry, grief, despair, depressed \\
Good things happening & joy, contented, pleased, delighted, excited\\
People doing bad thing & anger, indignation, shocked, appalled, hurt\\
Thinking about ourselves & remorse, guilt, shame, humiliation, embarrassment, pride, triumph\\
Unclassified & frustration, relief, disappointment, surprise, amazement, happy (gl{\"u}cklich, heureux, schastlivyi), frightened\\
\bottomrule
\end{tabularx}
\end{table*}

\begin{table*}[ht]
\centering
\caption{Emotion concepts associated to cognitive templates as analyzed by~\citet{wierzbicka1999}.}
\label{tbl:tem}
\begin{tabularx}{\textwidth}{lp{12.2cm}}
\toprule
Template & Emotion concepts \\
\midrule
First-person thought-plus-feeling & (+) great, wonderful, terrific, awesome, fabulous (--) awful, dreadful, terrible \\
Experiential evaluation & (+) entertaining, delightful, fascinating, compelling, interesting, touching (--) boring, predictable\\
Experiential with bodily reaction & (+) gripping, exciting, stunning, suspenseful, tense (--) disgusting, sickening\\
Lasting effect & (+) powerful, memorable, haunting, inspiring (--) depressing, disturbing\\
Cognitive evaluation & (+) complex, excellent, outstanding, impressive, brilliant, clever, original (--) disappointing, dismal, woeful\\
\bottomrule
\end{tabularx}
\end{table*}

Table~\ref{tab:rules-full} lists all thirteen rules of the mapper $g$ as a priority-ordered decision list (index $0$ highest). The first rule whose conditions hold fires; no match yields \textsc{abstain}. Each condition is transcribed from a published NSM explication; none is learned.

\begin{table}[h]
\centering
\setlength{\tabcolsep}{3.5pt}
\resizebox{\columnwidth}{!}{%
\begin{tabular}{cll}
\toprule
\# & Emotion & Slot conditions (conjunction) \\
\midrule
0  & relief     & WANT=\sv{not-want}, REALIZATION=\sv{not-happened}, EVALUATION=\sv{feel-good} \\
1  & surprise   & KNOWLEDGE=\sv{not-know}, REALIZATION=\sv{happened} \\
2  & fear       & EVALUATION=\sv{feel-bad}, TIME-DIR=\sv{after-now}, REALIZATION=\sv{maybe} \\
3  & boredom    & TRIGGER=\sv{nothing}, WANT$\in\{$\sv{want},\sv{not-want}$\}$, EVALUATION=\sv{feel-bad} \\
4  & disgust    & EVALUATION=\sv{feel-bad}, BODY=\sv{yes}, WANT=\sv{not-want}, EVAL-TARGET$\in\{$\sv{object},\sv{other}$\}$ \\
5  & shame      & EVALUATION=\sv{feel-bad}, EVAL-TARGET=\sv{self}, OTHERS-KNOW=\sv{can-know} \\
6  & guilt      & EVALUATION=\sv{feel-bad}, EVAL-TARGET=\sv{self}, AGENCY=\sv{i} \\
7  & pride      & EVALUATION=\sv{feel-good}, EVAL-TARGET=\sv{self}, AGENCY=\sv{i} \\
8  & trust      & EVALUATION=\sv{feel-good}, EVAL-TARGET=\sv{other}, KNOWLEDGE=\sv{know} \\
9  & anger      & EVALUATION=\sv{feel-bad}, AGENCY=\sv{someone-else}, EVAL-TARGET=\sv{other} \\
10 & sadness    & EVALUATION=\sv{feel-bad}, REALIZATION=\sv{not-can} \\
11 & joy        & EVALUATION=\sv{feel-good}, TRIGGER$\in\{$\sv{something-happened},\sv{someone-did}$\}$ \\
12 & no-emotion & EVALUATION=\sv{neither} \\
\bottomrule
\end{tabular}}
\caption{The complete mapper $g$. Shame precedes guilt (both share \sv{feel-bad}$\wedge$\sv{self}, split by OTHERS-KNOW vs.\ AGENCY); the general joy rule is last so specific positive emotions shadow it.}
\label{tab:rules-full}
\end{table}

\section{Per-Slot Annotation Reliability}
\label{app:perslot}
Table~\ref{tab:perslot} gives the full per-slot raw agreement and Krippendorff's $\alpha$ for the silver pass and the first human pass.

\begin{table}[h]
\centering\small
\begin{tabular}{lccc}
\toprule
Slot & Agree.\ (S) & $\alpha$ (S) & $\alpha$ (H) \\
\midrule
EXPERIENCER    & 0.994 & 0.00  & $-0.01$ \\
EVALUATION     & 0.865 & 0.768 & 0.227 \\
AGENCY         & 0.827 & 0.737 & 0.607 \\
WANT           & 0.821 & 0.696 & 0.395 \\
EVAL-TARGET    & 0.673 & 0.548 & 0.702 \\
REALIZATION    & 0.744 & 0.542 & 0.788 \\
TRIGGER        & 0.705 & 0.495 & 0.656 \\
TIME-DIRECTION & 0.686 & 0.449 & 0.192 \\
OTHERS-KNOW    & 0.885 & 0.437 & 0.316 \\
INTENSITY      & 0.788 & 0.425 & 0.230 \\
BODY           & 0.885 & 0.408 & 0.402 \\
KNOWLEDGE      & 0.769 & 0.391 & 0.081 \\
\midrule
Mean           & 0.803 & 0.491 & 0.382 \\
\bottomrule
\end{tabular}
\caption{Per-slot raw agreement and Krippendorff's $\alpha$. S = silver model pass ($n{=}156$); H = first human pass ($n{=}99$). High raw agreement co-occurs with low $\alpha$ on skewed slots (EXPERIENCER, OTHERS-KNOW, BODY), so the chance-corrected coefficient is the one to read.}
\label{tab:perslot}
\end{table}

\section{Mapper Behaviour: Typed Abstention, Priority Sensitivity, Distributional Evaluation, and Confidence Intervals}
\label{app:mapper-details}
This appendix reports four diagnostics of the mapper $g$ that the main text summarizes: the composition of abstention by type (Table~\ref{tab:abstain}), the sensitivity of accuracy to rule ordering (Table~\ref{tab:priority}), a distributional scoring of $g$ as a 13-way label distribution (Table~\ref{tab:dist}), and exact binomial confidence intervals on the held-out set (Table~\ref{tab:ci}).

\begin{table}[h]
\centering\small
\begin{tabular}{lrr}
\toprule
Abstention type & Full batch & 36-eval \\
\midrule
No-rule (legal, no rule fires)   & 141 (11.8\%) & 12 \\
Schema-residue (residue present) & 25 (2.1\%)   & 0  \\
Illegal (not schema-legal)       & 0            & 0  \\
\midrule
Total abstain                    & 166 (13.8\%) & 12 (33.3\%) \\
Multi-rule (resolved by priority)& 342 (28.5\%) & --- \\
\bottomrule
\end{tabular}
\caption{Typed abstention (A). Abstention is dominated by no-rule-fires, not malformed explications, and multi-rule items never abstain because priority resolves every tie.}
\label{tab:abstain}
\end{table}

\begin{table}[h]
\centering\small
\begin{tabular}{@{}lc@{}}
\toprule
Mapper configuration & Full-batch acc \\
\midrule
NSM most-specific-first (ours)   & 0.382 \\
Swap: guilt$\leftrightarrow$shame & 0.388 \\
Random orders ($N{=}200$): min/mean/max & 0.295 / 0.343 / 0.392 \\
\midrule
Single-winner selective acc      & 0.443 \\
Set-valued (gold $\in$ fired set) & 0.521 \\
\bottomrule
\end{tabular}
\caption{Priority-order sensitivity (A). Our hand order sits at the $96$th percentile of random orders; the guilt/shame swap moves accuracy under a point. Set-valued accuracy is $0.521$ vs.\ $0.443$ single-winner: about eight points live in priority resolution.}
\label{tab:priority}
\end{table}

\begin{table}[h]
\centering\small
\setlength{\tabcolsep}{4pt}
\begin{tabular}{lrrrrr}
\toprule
Scoring rule    & NLL  & Brier & top-1 & top-3 & cov.\ \\
\midrule
single-winner   & 2.206 & 0.833 & 0.398 & 0.498 & 0.862 \\
uniform-fired   & 2.140 & 0.826 & 0.328 & 0.558 & 0.862 \\
priority-decay  & \textbf{2.126} & \textbf{0.817} & \textbf{0.398} & \textbf{0.558} & 0.862 \\
\bottomrule
\end{tabular}
\caption{Distributional evaluation of $g$ on tier A ($n{=}1{,}200$) with Laplace smoothing ($\alpha{=}0.05$). Priority-decay strictly dominates single-winner: top-3 recovers six points, NLL and Brier both fall. Set-valued accuracy (gold $\in$ fired set) is $0.449$.}
\label{tab:dist}
\end{table}

\begin{table}[h]
\centering\small
\begin{tabular}{lcc}
\toprule
System ($n{=}36$) & Acc & 95\% CI \\
\midrule
Fine-tuned 8B (ours)      & 0.333 & $[0.19,\,0.51]$ \\
Black-box 8B              & 0.389 & $[0.23,\,0.57]$ \\
Appraisal (oracle)        & 0.361 & $[0.21,\,0.54]$ \\
Gold explication (routed) & 0.306 & $[0.16,\,0.48]$ \\
\bottomrule
\end{tabular}
\caption{Exact (Clopper--Pearson) $95\%$ binomial intervals on the held-out set. All four overlap, so no pairwise accuracy difference is resolvable at $n{=}36$; we read Table~\ref{tab:e2main} as a trade-off, not a ranking.}
\label{tab:ci}
\end{table}

\section{Additional Result Figures}
\label{app:verifier}
\begin{figure}[h]
\centering
\includegraphics[width=\columnwidth]{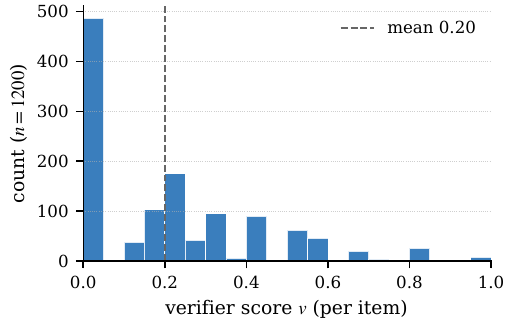}
\caption{Per-item entailment verification score on the full batch (A, $n{=}1{,}200$). The verifier is conservative on prime-vocabulary propositions: $40.6\%$ of items score zero and the mean is $0.20$. On the silver subset the uncalibrated score separates high-divergence items at AUC $0.477$ and silver-correct items at AUC $0.576$, significant on neither (S).}
\label{fig:nli}
\end{figure}

\begin{figure*}[h]
\centering
\includegraphics[width=0.9\textwidth]{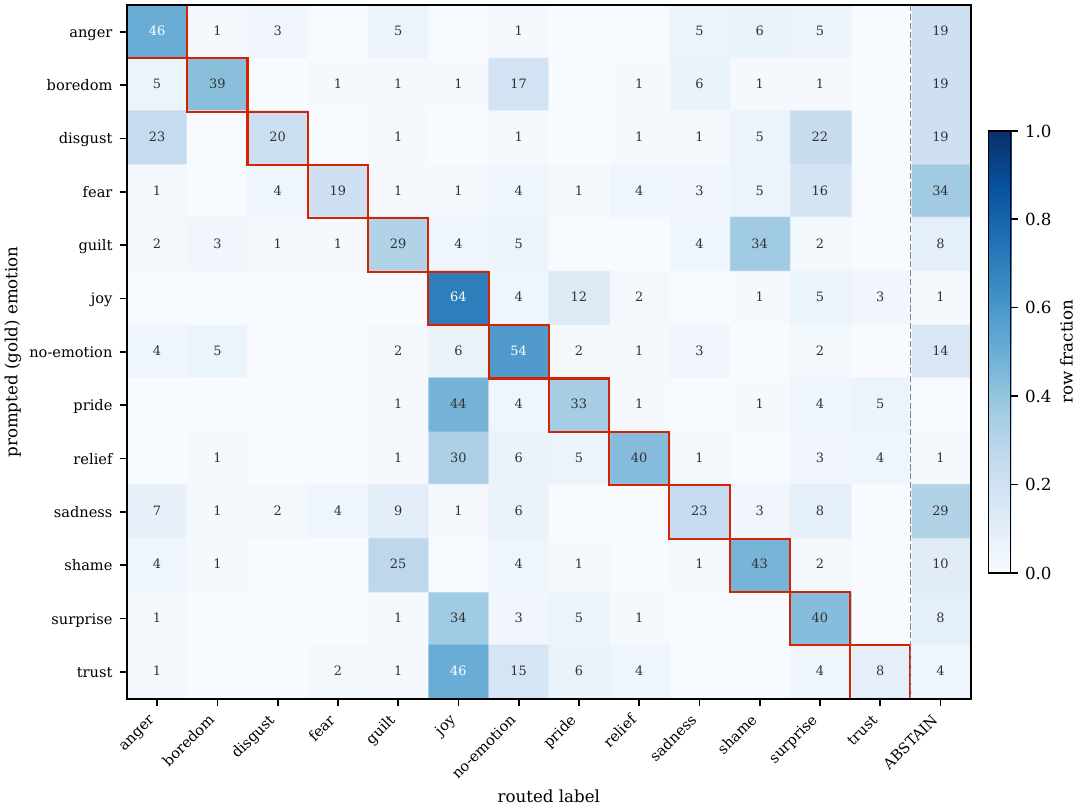}
\caption{Row-normalized routing confusion matrix on the full batch (A, $n{=}1{,}200$); raw counts shown, the diagonal is boxed and the right column is abstention.}
\label{fig:confusion}
\end{figure*}

This appendix collects the two full-resolution result figures summarized in Section~\ref{sec:results}: the per-item verification histogram (Figure~\ref{fig:nli}) and the complete routing confusion matrix (Figure~\ref{fig:confusion}).

\end{document}